\documentclass{bvm}
\begin{document}
\selectlanguage{english}

\title{How Robust Are Prognostic Radiomic Features?}
\subtitle{Comparing Selection Techniques on Pooled Glioblastoma Cohorts}
\titlerunning{How robust are prognostic radiomic features?}

% ===================== AUTHOR BLOCK =====================
% For the double-blind REVIEW submission comment out \author, \authorrunning,
% \institute and \email (BVM guidelines, Sec. 4.1). 0000_review.pdf is built that way.
% TODO: all entries marked TODO must be checked/filled by the authors.
\author{
	\fname{Yannick} \lname{Sodonougbo} \inst{1} \affiliation{University of Augsburg} \authorsEmail{TODO@uni-a.de} \street{Universitätsstraße} \housenumber{2} \zipcode{86159} \city{Augsburg} \country{Germany},
	\fname{Marlene} \lname{Schnieders} \inst{1} \affiliation{University of Augsburg} \authorsEmail{TODO@uni-a.de} \street{Universitätsstraße} \housenumber{2} \zipcode{86159} \city{Augsburg} \country{Germany},
	\fname{Alexia} \lname[0009-0005-6536-3122]{Rizoudis} \inst{1} \affiliation{University of Augsburg} \authorsEmail{TODO@uni-a.de} \street{Universitätsstraße} \housenumber{2} \zipcode{86159} \city{Augsburg} \country{Germany},
	\fname{Santiago} \lname[0000-0003-1667-8548]{Cepeda} \inst{2,3} \affiliation{Río Hortega University Hospital} \authorsEmail{TODO} \street{TODO} \housenumber{TODO} \zipcode{TODO} \city{Valladolid} \country{Spain},
	\fname{Frank} \lname[0000-0002-2857-7122]{Kramer} \inst{1} \affiliation{University of Augsburg} \authorsEmail{TODO@uni-a.de} \street{Universitätsstraße} \housenumber{2} \zipcode{86159} \city{Augsburg} \country{Germany},
	\fname{Dominik} \lname[0000-0003-0838-9885]{Müller} \inst{1,4} \affiliation{Ulm University of Applied Sciences} \authorsEmail{dominik.mueller@thu.de} \street{Prittwitzstraße} \housenumber{10} \zipcode{89075} \city{Ulm} \country{Germany} \isResponsibleAuthor
}
\authorrunning{Sodonougbo et al.}
\institute{
\inst{1} IT Infrastructure for Translational Medical Research, University of Augsburg\\
\inst{2} Department of Neurosurgery, Río Hortega University Hospital\\
\inst{3} Specialized Group in Biomedical Imaging and Computational Analysis, Instituto de Investigación Biosanitaria de Valladolid\\
\inst{4} Department of Computer Science, Ulm University of Applied Sciences
}
\email{dominik.mueller@thu.de}
% =========================================================

\maketitle

\begin{abstract}
Radiomic survival studies in glioblastoma report heterogeneous prognostic features, but how much of this stems from the methodology itself is rarely examined. We analysed the association of 463 radiomic, anatomical and clinical features from pre-treatment MRI with overall survival in 705 IDH-wildtype glioblastoma patients from three public cohorts, using three techniques: a log-rank test after median split, univariate Cox regression (both Benjamini--Hochberg corrected) and an XGBoost regressor interpreted with SHAP. The log-rank test selected 99 features, Cox regression 72 and SHAP its top 20; five were selected by all three: age, a peritumoural T2 texture feature, the number of connected peritumoural mask components and two image-level intensity maxima. The two survival tests shared 51 features, about half of the SHAP top 20 were also selected by either test, and where selections overlapped, effect directions agreed. Many selections were PyRadiomics diagnostic descriptors that are not restricted to tumour voxels. Analytical pipeline choices, including ranking method, pre-filtering and descriptor handling, can substantially influence which features appear prognostic.
\end{abstract}

\section{Introduction}
Despite maximal resection and chemoradiotherapy, median overall survival (OS) of glioblastoma is about 15 months~\cite{0000-stupp}. Radiomics converts pre-treatment MRI into high-dimensional quantitative descriptors of tumour shape, intensity and texture, and studies have reported radiomic signatures with prognostic value~\cite{0000-kickingereder,0000-prasanna}.

The reported features, however, differ considerably between studies, and besides cohorts and extraction settings, the ranking technique itself may influence which features are found. Classical survival tests such as the log-rank test and Cox regression account for censoring and test one feature at a time, whereas gradient-boosted trees model non-linear interactions and are ranked post hoc with explanation methods such as SHAP~\cite{0000-xgboost,0000-shap}. In addition, radiomics toolkits emit technical descriptors alongside the actual features~\cite{0000-pyradiomics}; if these are not removed, they may be mistaken for biomarkers.

We therefore compared three feature-ranking techniques on a multi-cohort glioblastoma dataset and asked (i) how strongly the selected feature sets agree, (ii) which features are consistently selected, and (iii) which share of the selections reflects technical rather than tumour-related information.

\section{Materials and methods}
\subsection{Data and features}
We included 705 patients with IDH-wildtype glioblastoma, available age and OS from three public cohorts, RHUH-GBM ($n=36$)~\cite{0000-cepeda}, UPenn-GBM ($n=309$)~\cite{0000-bakas} and UCSF-PDGM ($n=360$)~\cite{0000-calabrese}, after excluding ten patients with missing sub-region labels. Death was observed in 554 patients; 151 were right-censored; survival status was harmonised from cohort-specific encodings. Images were co-registered, skull-stripped, z-score normalised and bias-field corrected. Reference segmentations, provided by the RHUH collaborators or with the public datasets, distinguish enhancing tumour (ET), necrotic core (NC) and the T2/FLAIR-hyperintense peritumoural zone (PERI); total tumour (TT) denotes their union, i.e.\ the BraTS whole tumour. Radiomic features were extracted with PyRadiomics~\cite{0000-pyradiomics} from native (non-contrast) T1-weighted images for ET and NC and from T2-weighted images for PERI, on the original image and the wavelet sub-bands LL, LH, HL and HH, each with and without intensity normalisation. The radiomic descriptors were restricted to ten texture and shape features that an earlier analysis within our collaboration had selected by their correlation with OS across all sequences and sub-regions, on data overlapping with the present cohort; the sequence per sub-region followed this selection. They were complemented by sub-region volumes, centroid deviations, overlaps with the Harvard--Oxford atlases after affine registration, and age and sex. After removing constant, tuple-valued and identifier columns, 463 features remained.

The feature table also contained PyRadiomics \emph{diagnostics} output~\cite{0000-pyradiomics}. Mask-derived quantities (voxel count, i.e.\ volume; number of connected mask components; mean intensity within the mask) describe the segmented region. Image-level statistics (minimum, maximum, mean) are not restricted to tumour voxels: the original ones cover the entire image, the interpolated ones the mask bounding box plus padding.

\subsection{Feature-ranking techniques}
\paragraph{Log-rank test} Each feature was split at its median (values $\ge$ median formed the high group), and the two groups were compared with the log-rank test. Features whose split produced a group with fewer than ten patients were excluded (240 features, all atlas overlaps with mostly zero values), leaving 223 tested features.
\paragraph{Cox regression} Features with variance below 0.01 on the unstandardised scale were excluded, leaving 230 tested features; the remaining features were z-standardised and entered into univariate Cox proportional hazards models, so that the hazard ratio (HR) refers to one standard deviation (age: 11.8 years). For both survival tests, $p$-values were corrected with the Benjamini--Hochberg procedure~\cite{0000-bh} and features with $q \le 0.05$ were selected.
\paragraph{XGBoost with SHAP} An XGBoost regressor~\cite{0000-xgboost} with absolute-error objective was trained on all 463 features to predict OS in days (learning rate 0.05, L1 and L2 regularisation 15, subsample 0.8, column subsample 0.9, early stopping after 100 rounds), using five-fold cross-validation stratified by survival group (OS $<170$, 170--599, $>599$ days), balanced sample weights and iterative imputation. This regression did not model censoring; censored times were treated as observed. Importance was the mean absolute SHAP value over all test folds~\cite{0000-shap}, and the 20 highest-ranked features were selected; this cut-off was chosen pragmatically, as the SHAP values declined smoothly without a natural threshold.

\subsection{Comparison}
Selections were compared at feature level. For visualisation only, features that differ only by wavelet sub-band, normalisation or interpolation were grouped into families. Effect directions were taken from the median OS of the high and low group (log-rank), the sign of $\log$ HR (Cox) and the Spearman correlation between feature value and SHAP value over all test folds (SHAP). External evidence for each family was contextualised qualitatively by a targeted, non-systematic literature search.

\section{Results}
\begin{figure}[!b]
\centering
\includegraphics[width=\textwidth]{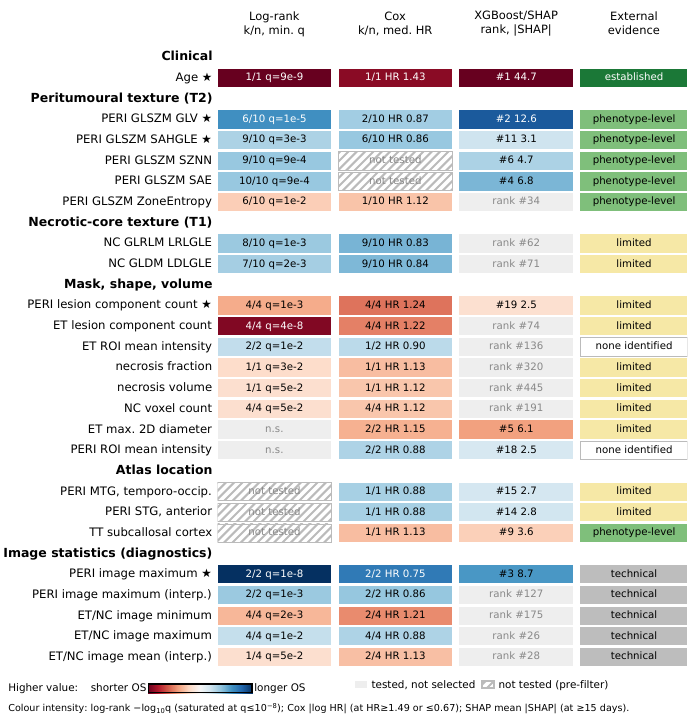}
\caption{Feature families selected by at least two techniques. Cells: selected/all variants ($k/n$), smallest $q$, median HR of the selected variants, best SHAP rank and mean $|$SHAP$|$ (days). GLV, gray-level variance; SAHGLE, small-area high-gray-level emphasis; SZNN, normalised size-zone non-uniformity; SAE, small-area emphasis; LRLGLE, long-run low-gray-level emphasis; LDLGLE, large-dependence low-gray-level emphasis. $\star$: family selected by all three techniques, not necessarily in the same variant. External evidence (qualitative): established, recognised clinical prognostic factor; phenotype-level, prior evidence for the compartment or imaging phenotype but not the exact descriptor; limited, indirect or inconsistent evidence; none identified; technical, image-level statistic not restricted to tumour voxels.}
\label{0000-fig-heatmap}
\altText{Table-like heatmap with one row per feature family, grouped into clinical, peritumoural texture, necrotic-core texture, mask/shape/volume, atlas location and image statistics, and four columns: log-rank test, Cox regression, SHAP and external evidence. Red cells indicate that higher values are associated with shorter survival, blue cells with longer survival; rows show the same colour across all techniques that selected them. Age is the only family with established external evidence; image statistics are marked as technical.}
\end{figure}

The three techniques selected sets of very different size and composition: the log-rank test 99 of 223 tested features, Cox regression 72 of 230 and SHAP 20 of 463. Texture features dominated both survival tests (61 and 27 selections), atlas location features made up 6 of the SHAP top 20, and image-level diagnostics accounted for 17, 15 and 2 selections, respectively. Five features were selected by all three techniques: age (Cox HR 1.43, 95\,\% CI 1.31--1.57), the peritumoural wavelet GLSZM gray-level variance (HR 0.87, 0.79--0.95), the number of connected peritumoural mask components (HR 1.24, 1.16--1.34) and the T2 image maximum with and without normalisation (HR 0.75, 0.68--0.83). At family level, the peritumoural GLSZM SAHGLE family was additionally selected by all three techniques, in different variants (Fig.~\ref{0000-fig-heatmap}).

Log-rank and Cox shared 51 features; about half of the SHAP top 20 were also selected by either survival test (log-rank 11, Cox 10). As SHAP was summarised by a fixed top 20 and the survival tests by an FDR threshold on different eligible feature sets, these counts are descriptive rather than comparable selection rates. Effect directions agreed in all 24 families selected by at least two techniques (Fig.~\ref{0000-fig-heatmap}); agreement between the survival tests is expected, as both assess the same marginal association, whereas agreement with SHAP is informative. Two texture groups were selected by both survival tests across nearly all filter variants: peritumoural GLSZM features of small zones and gray-level variance, and necrotic-core features indicating large homogeneous dark areas within the necrosis (LRLGLE, LDLGLE; median HR 0.83--0.84). Higher values were associated with longer OS in both groups, whereas higher peritumoural zone entropy was associated with shorter OS. Two peritumoural GLSZM families selected by the log-rank test in 9 and 10 of 10 variants were not tested by Cox regression, because the variance filter removed them.

The XGBoost model predicted OS with a mean absolute error of 243.7\,$\pm$\,10.9 days across folds, with predictions compressed to 100--700 days.

\section{Discussion}
The two censoring-aware survival tests selected largely the same features; as the median-split log-rank test addresses the same univariate association as the Cox model with less information~\cite{0000-altman}, it is not an independent second view. Some apparent disagreement reflects feature eligibility rather than conflicting evidence, as for the peritumoural families removed by the Cox variance filter. The lower overlap of the SHAP ranking may reflect differences in modelling objective, censoring treatment, feature eligibility and handling of correlated variants. Given the model's limited accuracy, its SHAP importance should not be interpreted as independent biomarker evidence.

Age provided an internal positive control: all three techniques selected this established prognostic factor. Beyond age, external support exists mainly at the level of imaging phenotypes rather than individual descriptors (Fig.~\ref{0000-fig-heatmap}), consistent with a meta-analysis by Choi et al.\ that reported a pooled C-index of 0.65 and high between-study heterogeneity for radiomic survival models in glioblastoma~\cite{0000-choi}. The peritumoural texture findings are broadly consistent with reports that the peritumoural zone carries prognostic information~\cite{0000-prasanna}, and the adverse subcallosal effect with the poorer survival of central tumours in a population-based atlas~\cite{0000-fyllingen}; the favourable temporal effects, however, contrast with the shorter survival reported there for left temporal-pole tumours, which our non-lateralised cortical atlas cannot resolve. The component count may partly reflect lesion multiplicity, which is associated with poorer outcome~\cite{0000-dicarlo}, but also segmentation fragmentation. None of these studies validated the specific descriptors selected here. The texture directions do not support a simple reading as heterogeneity: zone entropy, unlike small-zone and gray-level-variance features, was associated with shorter OS, and homogeneous dark necrotic areas with longer OS although necrosis volume was adverse. Texture features also depend on region size~\cite{0000-welch} and on acquisition and preprocessing~\cite{0000-traverso}.

Many selections were PyRadiomics diagnostics. Mask-derived quantities are interpretable, e.g.\ the voxel count is the region volume, whereas image-level statistics are not tumour-specific; accordingly, the original image statistics of ET and NC, computed on the same T1 image, yielded identical hazard ratios. The T2 image maxima acted as an informal negative control: consistent selection by all techniques did not imply tumour-specific relevance, suggesting that such descriptors may act as proxies for cohort, scanner, protocol or patient characteristics. Image-level diagnostics should be excluded from candidate biomarker sets, and mask-derived ones treated explicitly as morphological variables.

Several limitations apply. The radiomic descriptors had been pre-selected for their correlation with OS on overlapping data, which likely inflates their associations. The cohorts were pooled, so robustness across cohorts was not assessed; all analyses were univariate and unadjusted for cohort, treatment and region volume, and the proportional hazards assumption was not assessed; cohorts may differ in acquisition, segmentation, follow-up and outcome definitions. With a SHAP top 10, 30 or 50 instead of 20, 2, 8 or 9 features would have been selected by all three techniques. The XGBoost model did not model censoring, its stratification and weights used censored times, and early stopping and imputation involved the evaluation folds, so its error is optimistically biased.

In conclusion, which features are identified as prognostic depends strongly on analytical pipeline choices, including feature filtering, descriptor handling and ranking strategy.

\printbibliography
\end{document}